\documentclass[letterpaper, 10 pt, conference]{ieeeconf}  

\IEEEoverridecommandlockouts                              

\title{\LARGE \bf
PanoImager: Geometry-Guided Novel View Synthesis and Reconstruction from Sparse Panoramic Views
}

\author{Zhisong~Xu$^{1}$ and Takeshi~Oishi$^{1}$%
\thanks{$^{1}$Zhisong XU and Takeshi Oishi are with the Institute of Industrial Science,
The University of Tokyo, Tokyo, Japan
{\tt\small \{zhisongxv, oishi\}@cvl.iis.u-tokyo.ac.jp}}%
}

\usepackage{graphicx}
\usepackage{amsmath}
\usepackage{amssymb}
\usepackage{amsfonts}
\usepackage{booktabs}
\usepackage{makecell}

\usepackage[T1]{fontenc}
\usepackage[utf8]{inputenc} 
\usepackage{cite}

\begin{document}

\maketitle
\thispagestyle{empty}
\pagestyle{empty}

\begin{abstract}
Panoramic sensing offers wide field-of-view coverage, yet 3D reconstruction from sparse panoramas remains challenging under rotation-dominant, weak-parallax motion. 
In such regimes, SfM/SLAM initialization is often ill-conditioned and unreliable. 
We present \emph{PanoImager}, an SfM-free framework that combines feed-forward pose/depth priors, geometry-conditioned diffusion view completion, and depth-guided 3DGS optimization. 
Given only a few panoramic images, PanoImager decomposes them into local perspective views, synthesizes auxiliary observations to enrich sparse evidence, and stabilizes Gaussian optimization for improved cross-view consistency. 
Experiments on multiple benchmarks show improved stability under extreme sparsity, suggesting PanoImager as an offline/background component for map refinement when SfM/SLAM fails to initialize.

\end{abstract}
\section{Introduction}

\begin{figure*}[t]
  \centering
  \includegraphics[width=\textwidth]{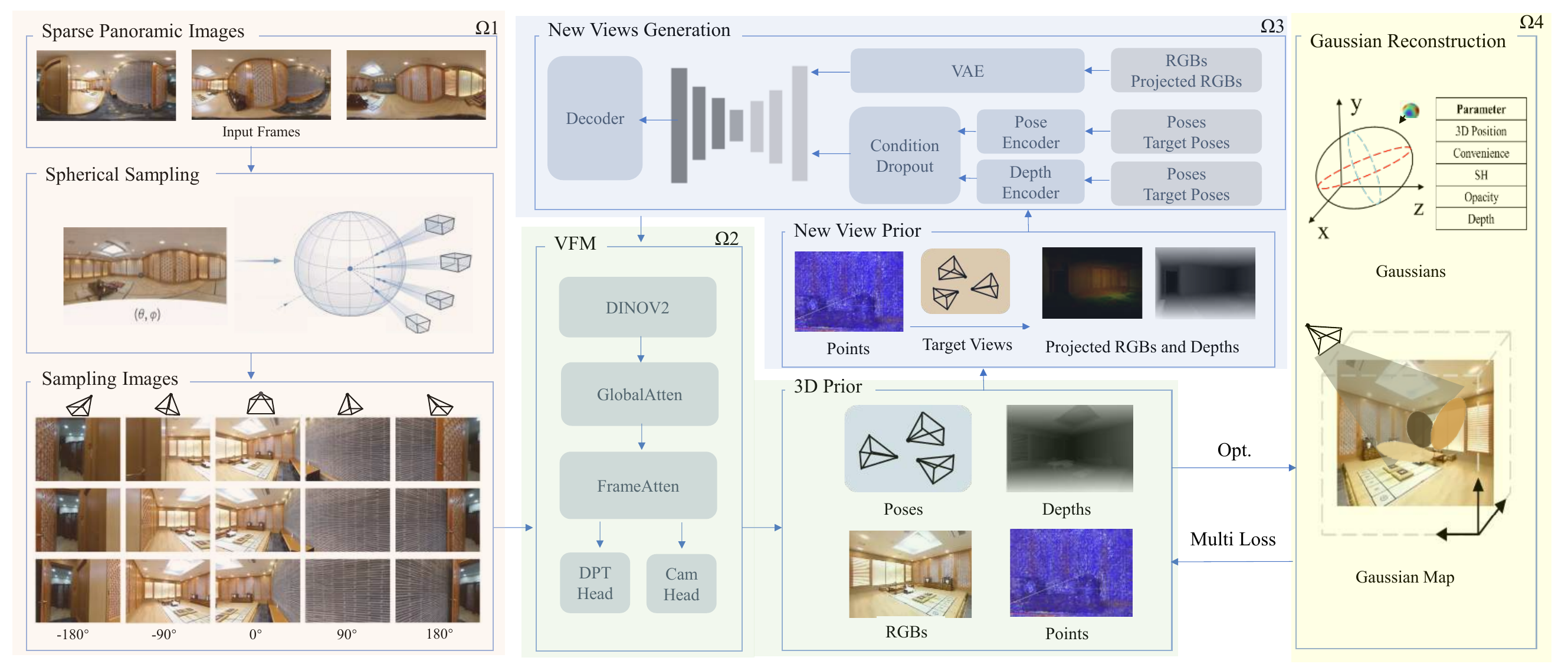}
  \caption{\textbf{Overview of our panoramic reconstruction pipeline.}
Starting from sparse panoramic images in $\Omega_{1}$, we first sample them into a perspective observation set in $\Omega_{2}$.
A visual foundation model then predicts camera poses and depth on $\Omega_{2}$, providing geometric priors.
Conditioned on these priors, we extend the observations from $\Omega_{2}$ to $\Omega_{3}$ via novel view synthesis, and map the generated views back to $\Omega_{2}$ to enrich the observation set.
Finally, the augmented observations are lifted to $\Omega_{4}$, where a 3D Gaussian representation is optimized for reconstruction and rendering.
    }
  \label{fig:pipeline}
\end{figure*}
Three-dimensional (3D) reconstruction plays a central role in robotic perception by supporting spatial reasoning, mapping, and interaction in complex environments~\cite{argyros2005robot}. In practical settings such as indoor inspection, teleoperation, and narrow-space exploration, robots often observe the scene under constrained motion and limited sensing budgets. In particular, rapid in-place scanning often leads to rotation-dominant capture with only weak translational parallax, making reliable reconstruction from sparse observations fundamentally difficult.

This limitation is not merely an engineering issue, but a geometric one. Conventional SfM and SLAM pipelines rely on sufficient viewpoint diversity and parallax to triangulate stable 3D structure. Under pure rotation or extremely small baselines, triangulation becomes unstable, geometric constraints become poorly conditioned, and initialization may become effectively rank-deficient. As a result, feature-correspondence-based methods often fail to provide a reliable starting point for downstream mapping and reconstruction.

Panoramic cameras are attractive for robotic mapping and localization because they capture a full $360^\circ$ field of view in a single image~\cite{chen2024panoslam,chen2024360orb,zhou2023inf,lo2017360}. While this wide coverage improves scene observability under sparse capture, panoramic imagery also introduces severe spherical distortion and large appearance variation, which complicate both geometric estimation and learning-based reconstruction. Thus, panoramic sensing alleviates field-of-view limitations, but does not by itself resolve the geometric degeneracy caused by weak-baseline motion.

Recent neural scene representations, such as NeRF~\cite{mildenhall2021nerf} and 3D Gaussian Splatting (3DGS)~\cite{kerbl20233d}, have significantly improved novel-view synthesis and 3D reconstruction. However, in panoramic settings, they remain sensitive to projection distortion~\cite{lee2024odgs,choi2023balanced,li2025spags} and still depend on geometry initialization or multi-view optimization under sufficiently well-conditioned motion. 
However, with extremely sparse panoramic inputs, SfM/SLAM initialization becomes unreliable and frequently fails. Feed-forward geometric models can directly predict depth and camera pose from sparse inputs~\cite{wang2024dust3r,wang2025vggt,xu2026framevggt}, and recent panoramic variants further reduce dependence on iterative optimization~\cite{lee2025omnisplat,zhang2025pansplat,chen2025splatter}. Yet under extreme sparsity, they often produce incomplete geometry and view-dependent artifacts. Diffusion models offer a complementary prior for synthesizing plausible observations from limited inputs~\cite{blattmann2023stable,yu2024viewcrafter,liu2024reconx,cao2025mvgenmaster}, but diffusion alone cannot ensure cross-view geometric consistency.

To address these challenges, we propose \emph{PanoImager}, a prior-guided SfM-free framework for 3D reconstruction from sparse panoramic observations in degenerate robotic exploration regimes. Instead of treating generative completion as an independent view synthesis module, PanoImager uses it as a geometry-aware scaffold for reconstructing a consistent 3D Gaussian scene. Specifically, feed-forward pose and depth priors provide an initial geometric anchor, pose- and depth-guided diffusion densifies the sparse observation space, and depth-constrained 3DGS optimization integrates real and synthesized views through reliability-aware geometric supervision. This design removes the dependence on stable feature correspondences across sparse panoramas and improves reconstruction robustness under rotation-dominant, weak-parallax, or feature-poor capture.

Our contributions are summarized as follows:
\begin{itemize}
\item We present \emph{PanoImager}, an SfM-free reconstruction framework for extremely sparse panoramic observations under rotation-dominant, weak-parallax motion, where conventional SfM/SLAM initialization becomes poorly conditioned or fails altogether.
\item We propose a training-free, geometry-conditioned view augmentation scheme that synthesizes auxiliary observations to densify the sparse measurement space, while using reliability-aware soft supervision to mitigate hallucinated content.
\item We introduce a depth-constrained 3D Gaussian optimization strategy with explicit geometric anchors and anti-floater regularization to improve structural integrity and spatial consistency in the reconstructed scene.
\end{itemize}

\section{Related Work}

\subsection{Panoramic Gaussian Splatting}

3D Gaussian Splatting (3DGS) enables efficient, high-quality novel view synthesis using explicit geometric primitives~\cite{kerbl20233d}. While sphere-native extensions adapt 3DGS to panoramic imagery via spherical rays or tangent-plane parameterizations~\cite{lee2024odgs, choi2023balanced, li2025spags}, these modifications often make optimization sensitive under sparse observations. When only a few panoramas are available with rotation-dominant motion, limited overlap leaves large unobserved regions, degrading geometric completeness, cross-view consistency, and extrapolation to unseen viewpoints—motivating the incorporation of structured view completion priors while preserving the stable optimization of perspective 3DGS.
\subsection{SfM and Feedforward Geometry}

Conventional 3D reconstruction pipelines typically rely on SfM or SLAM frameworks to estimate camera poses and recover scene geometry, including large-scale multi-sensor mapping systems and recent dense Gaussian SLAM approaches~\cite{zhang2025yuto,cowley2021upslam,hu2025mgso,zhu2025fgo,mmdslam}.
However, for panoramic imagery with wide fields of view and predominantly rotational motion, pose estimation from sparse observations remains challenging.
Nevertheless, in the extremely sparse panoramic setting, predictions can still be uncertain in unobserved regions, and the lack of dense multi-view constraints makes it difficult to recover coherent geometry across the full viewing sphere, motivating geometry-aware view completion to densify observations.

\subsection{Diffusion-Guided View Synthesis}

Diffusion-based generative models have recently shown strong capability in synthesizing novel views from sparse observations by leveraging learned appearance and structural priors.
Recent approaches further incorporate explicit geometric cues (e.g., depth- or pose-conditioned warping priors) to improve multi-view consistency and enable more reliable 3D optimization.
Nevertheless, diffusion models are predominantly developed for perspective imagery and are rarely tailored to panoramic sensing, where spherical distortions, structured view parameterization, and rotation-dominant motion impose stricter geometric constraints and exacerbate extrapolation to non-visible viewpoints.

\section{Method}

\subsection{Problem Definition and Overview}

The input consists of only $N_p$ panoramic frames, typically $N_p \in [3,6]$, leaving many 3D surfaces weakly constrained due to limited baseline, occlusion, and insufficient viewpoint diversity.
We denote this sparse panoramic input domain by $\Omega_{1}$. 
Under such limited spatial coverage and viewpoint diversity, reconstruction becomes severely ill-conditioned: insufficient viewpoint overlap and weak parallax lead to unstable pose estimation, fragmented geometry, and unreliable multi-view optimization. 
These failure modes make purely geometric pipelines fragile and motivate the introduction of stronger geometric and generative priors.

To address this setting, we propose \emph{PanoImager}, illustrated in Fig.~\ref{fig:pipeline}. 
Starting from $\Omega_{1}$, we first decompose each panorama into a structured set of local perspective views, yielding an intermediate domain $\Omega_{2}$. 
On $\Omega_{2}$, we build a \emph{feed-forward geometric scaffold} by directly predicting camera poses and depth maps $\{P_{l,k}, D_{l,k}\}$, thereby avoiding fragile multi-view cost-volume optimization in the original panoramic domain. 
To compensate for missing observations, we then perform geometry-conditioned view completion: a diffusion model, guided by warped RGB/depth cues and visibility masks $(I^{\mathrm{warp}}_{r\rightarrow t}, D^{\mathrm{warp}}_{r\rightarrow t}, M_{r\rightarrow t})$, synthesizes auxiliary views in $\Omega_{3}$. 
These completed views are mapped back to $\Omega_{2}$ and used as additional geometric support for the final reconstruction stage. 
Finally, the captured and synthesized observations are jointly used to optimize a depth-constrained 3D Gaussian representation in $\Omega_{4}$, producing a spatially more coherent reconstruction under sparse panoramic input.

Formally, we optimize
\begin{equation}
\hat{G}=\arg\min_G \ \mathcal{L}(G;\mathcal{U}),
\end{equation}
where $\mathcal{U}=\mathcal{U}_{\text{cap}}\cup\mathcal{U}_{\text{syn}}$ collects captured observations and synthesized views, together with their camera parameters and depth signals.
Crucially, synthesized views are not treated as hard measurements. 
They enter the objective as reliability-weighted soft photometric and depth priors under sparse coverage, while captured views remain the primary supervision.

\subsection{Panoramic Geometry Initialization}

\begin{figure}[t]
  \centering
  \includegraphics[width=\linewidth]{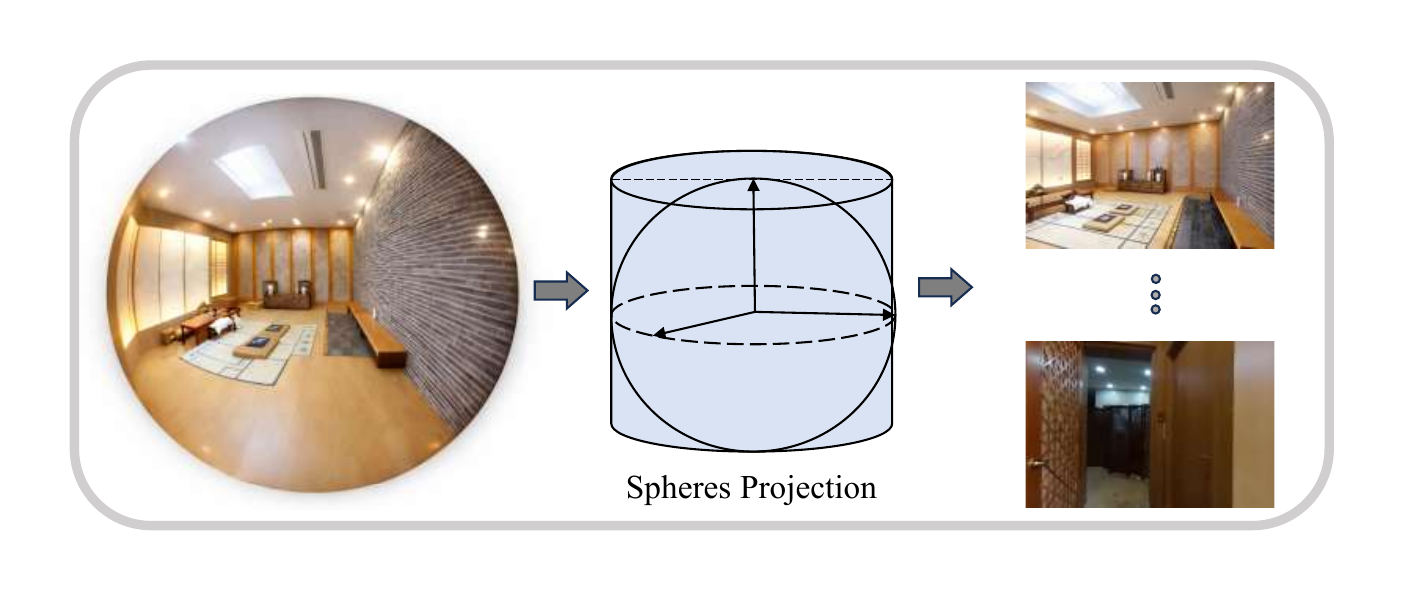}
  \caption{
    Each panorama is sampled over the viewing sphere and decomposed into a set of overlapping local perspective views, which approximate the panoramic camera with tangent-space perspective charts.}
  \label{fig:transition_plane}
\end{figure}

A central difficulty in panoramic reconstruction is that panoramic observations are naturally defined on the viewing sphere $\mathbb{S}^2$, while their equirectangular projection (ERP) provides a highly non-uniform 2D parameterization. 
Under ERP, equal pixel displacements do not correspond to equal angular displacements on the sphere; instead, the local projection scale varies with position and becomes increasingly anisotropic near high-latitude regions. 
As a result, small image-space perturbations can induce large directional errors, making direct pose estimation and geometry optimization in the ERP domain poorly conditioned under sparse capture.

To improve geometric conditioning, we re-parameterize each panorama in tangent space by decomposing it into a set of overlapping local perspective views,
\[
\mathcal{V}_l = \{(I_{l,k}, K_{l,k})\}_{k=1}^{N_v},
\]
obtained by quasi-uniform spherical sampling over $\mathbb{S}^2$. 
As shown in Fig.~\ref{fig:transition_plane}, the panorama is reformulated as a set of overlapping tangent-space perspective charts.
This decomposition converts a globally distorted panoramic parameterization into locally well-conditioned perspective charts that better match the pinhole camera model used by downstream geometric modules. 
In practice, we use $N_v=18$ viewing directions, which provides a good trade-off among spherical coverage, overlap between neighboring views, and computational cost. 
Fewer directions leave large regions weakly constrained, while substantially denser sampling mainly adds redundancy with limited geometric benefit.

This tangent-space reformulation is also important for stable 3D Gaussian Splatting (3DGS) optimization. 
Given a 3D Gaussian with covariance $\Sigma$, its projected covariance is approximated as
\[
\Sigma' = J \Sigma J^T,
\]
where $J$ denotes the local projection Jacobian. 
This first-order propagation is reliable only when the camera mapping remains locally smooth. 
In the ERP domain, however, the projection is strongly spatially varying, so the local Jacobian can change significantly even within a small neighborhood, reducing covariance accuracy and destabilizing optimization. 
By operating on normalized local perspective charts, the projection becomes more locally uniform, improving numerical conditioning for both geometric estimation and Gaussian optimization.

On this re-parameterized view set, we build an SfM-free, feed-forward geometry initializer based on the Visual Geometry Grounded Transformer (VGGT).
VGGT jointly encodes the perspective views $\mathcal{V}_l$ and aggregates geometric evidence across overlapping observations via cross-view attention, yielding aligned depth maps $\{D_{l,k}\}$ and relative poses $\{P_{l,k}\}$.
We also evaluated DUSt3R as an alternative initializer, but found it unreliable in the same sparse panoramic regime, exhibiting failure modes similar to SLAM/SfM initialization.
These predictions therefore provide a geometrically coherent scaffold without requiring stable feature correspondences, and initialize the subsequent view completion, observation densification, and depth-constrained Gaussian optimization stages.
\subsection{Geometry-Conditioned View Completion}

Reconstructing 3D scenes from sparse panoramic inputs is ill-posed due to limited coverage and weak geometric constraints, which often yields incomplete appearance support and unstable downstream optimization.
We therefore formulate observation densification as geometry-conditioned view completion, where auxiliary views are synthesized under the estimated pose/depth priors.
We freeze this geometric scaffold and densify only by sampling additional pose-aligned target cameras; importantly, the synthesized views are used as reliability-weighted soft priors to regularize optimization, acknowledging that they may influence geometry through the coupled photometric and depth objectives.

\begin{figure}[t]
  \centering
  \includegraphics[width=\linewidth]{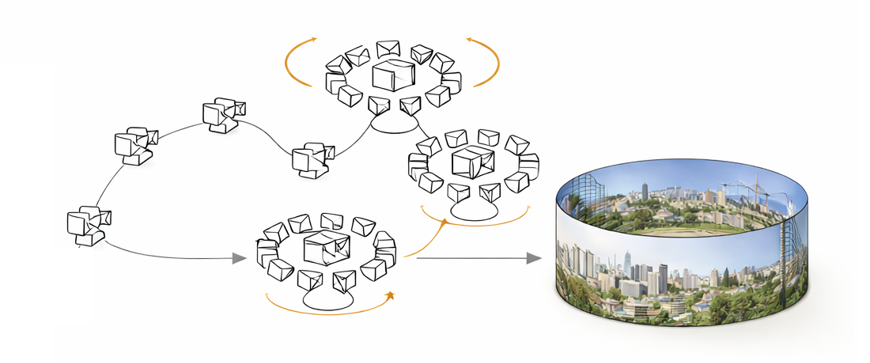}
  \caption{
  Panoramic view completion via geometry-guided sampling.
  A smooth trajectory is interpolated from the initialized poses, and $N{=}16$ local perspective targets are sampled uniformly along the spline.
  The synthesized views are later fused into a completed panoramic representation.
  }
  \label{fig:interpolation}
\end{figure}

\paragraph{Tangent-Space Completion.}
To bridge the domain gap between spherical Equirectangular Projection (ERP) and pinhole-optimized diffusion models, we perform completion in the tangent-space perspective domain.
By decomposing the panorama into local perspective charts, the diffusion process is conditioned on undistorted and physically meaningful geometric cues.
In this framework, the diffusion model acts as a geometrically-guided interpolaton (Fig.~\ref{fig:interpolation}) for missing observations, rather than an unconstrained generative engine.

\paragraph{Target Camera Sampling.}
We sample $N{=}16$ virtual target cameras $\mathcal{T}=\{(P_i,K_i)\}_{i=1}^{N}$ along a cubic spline fitted to the sparse input poses. Camera centers are distributed with \emph{equal arc-length} spacing and rotations are slerp-interpolated. All targets employ fixed pinhole intrinsics with $\mathrm{FoV}{=}90^\circ$ to balance tangent-chart distortion and spherical coverage. For each target $(P_t,K_t)$, we reproject reference observations to obtain warped RGB-D priors and visibility masks. To suppress occlusion-boundary artifacts, we discard pixels exceeding a local depth-discontinuity threshold:
\begin{equation}
\Delta D^{\mathrm{warp}}_{r\rightarrow t} > 0.05 \cdot d_{\max},
\end{equation}
where $\Delta D$ denotes the maximum depth jump within a 4-neighborhood and $d_{\max}$ is the maximum valid depth. These warped cues provide explicit geometric anchors, while the diffusion model infers appearance primarily in regions where direct evidence is sparse.
\paragraph{Geometry-Conditioned Diffusion.}
We adopt \textbf{MVGenMaster} as the diffusion backbone.
For a target view $t$ and a selected reference view $r\in\mathcal{R}(t)$, the denoiser predicts noise on the target latent $x_t^{(n)}$ conditioned on (i) the reference latent $x_r$, (ii) their camera parameters, and (iii) warped RGB-D priors and a validity mask:
\begin{equation}
\epsilon_\theta\!\left(
x_t^{(n)} \;\middle|\;
\begin{aligned}
&x_r,\ \big((P_r,K_r),(P_t,K_t)\big),\ I_r,\ I^{\mathrm{warp}}_{r\rightarrow t},\\
&\big(D_r, D^{\mathrm{warp}}_{r\rightarrow t}\big),\ M_{r\rightarrow t}
\end{aligned}
\right).
\end{equation}
Here $I_r\in\mathbb{R}^{H\times W\times 3}$ and $D_r\in\mathbb{R}^{H\times W}$ are the RGB image and predicted depth of reference view $r$ in the tangent-space perspective domain.
$P_r,P_t\in SE(3)$ denote the camera extrinsics (and $K_r,K_t$ the fixed pinhole intrinsics).
We obtain $I^{\mathrm{warp}}_{r\rightarrow t}$ and $D^{\mathrm{warp}}_{r\rightarrow t}$ by back-projecting $D_r$ to 3D in the $r$ frame and reprojecting to the $t$ image plane using $(P_r,K_r)$ and $(P_t,K_t)$.
The mask $M_{r\rightarrow t}\in\{0,1\}^{H\times W}$ marks pixels with valid reprojection after z-buffer and depth-edge filtering.
We concatenate/pool multi-reference cues, anchoring synthesis where $M_{r\rightarrow t}=1$.

\paragraph{Spherical Blending of Local Completions.}
The finalized panoramic representation is reconstructed by fusing the overlapping perspective latents back onto the spherical domain via normalized weighted blending.
Let $x_i$ denote the synthesized latent for target view $i$, and $\mathcal{T}^{-1}(\cdot)$ denote the inverse projection from the local chart to the sphere.
We define a confidence map $W_i$ using a 2D Gaussian kernel centered at the chart midpoint:
\begin{equation}
W_i(u,v) = \exp\!\left(-\frac{\|(u,v)-(u_0,v_0)\|_2^2}{2\sigma^2}\right),
\qquad \sigma = 0.25,
\end{equation}
where $(u,v)$ are normalized image coordinates and $(u_0,v_0)$ is the chart center.
The global spherical latent $\mathcal{L}_{\mathrm{sph}}$ is computed as:
\begin{equation}
\mathcal{L}_{\mathrm{sph}}
=
\frac{
\sum_{i=1}^{N}
\hat{M}_i \odot \hat{W}_i \odot \mathcal{T}^{-1}i(x_i)
}{
\sum{i=1}^{N} \hat{M}_i \odot \hat{W}_i + \epsilon
},
\end{equation}
where $\epsilon$ is a small constant for numerical stability. This blending suppresses seams between adjacent charts and emphasizes low-distortion central regions, yielding a coherent panoramic completion.

\begin{figure}[t]
  \centering
  \includegraphics[width=\linewidth]{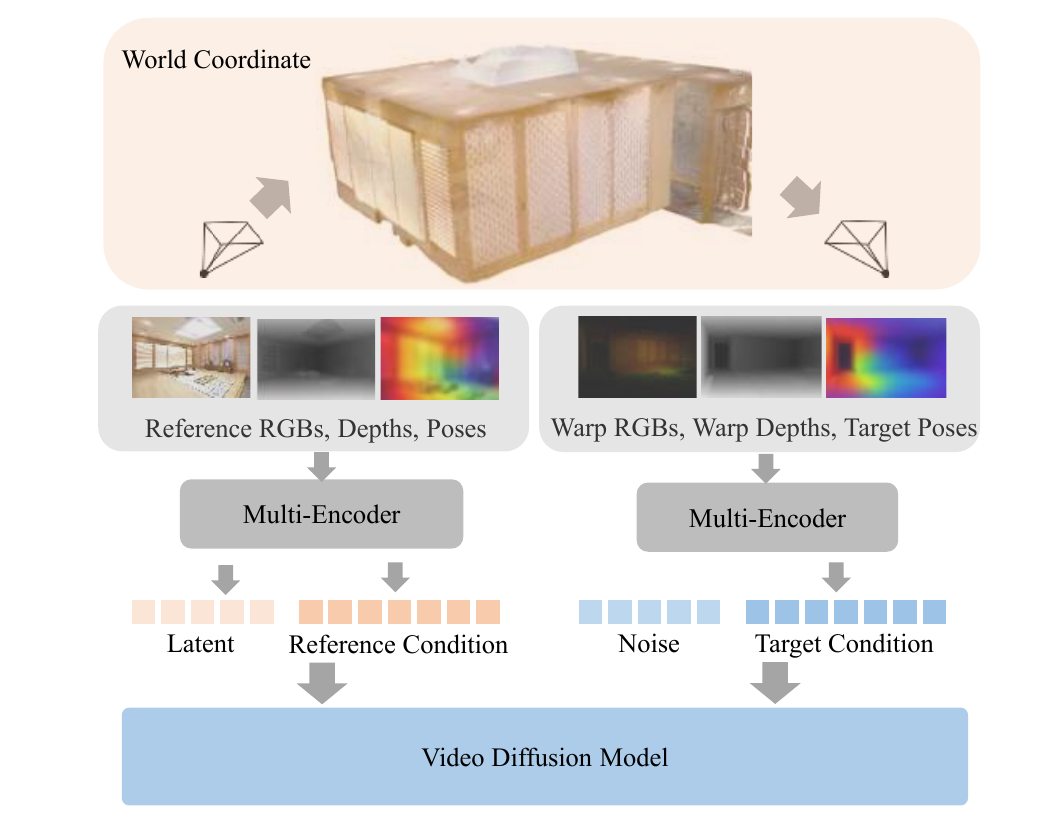}
  \caption{
  For each target view, reference observations are reprojected to construct warped RGB/depth priors and visibility masks, which serve as multi-modal geometric conditions for diffusion-based completion.
  The synthesized local views are fused on the spherical domain by Gaussian-weighted blending ($\sigma{=}0.25$).
  }
  \label{fig:diffusion}
\end{figure}

\paragraph{Reliability-Aware Condition Dropout.}
We gate target-side conditions $c_i={c_{\text{rgb}},c_{\text{pose}},c_{\text{depth}}}$ with $z_i\sim\mathrm{Bernoulli}(\omega_i)$:
\begin{equation}
\tilde{c}_i= z_i,c_i + (1-z_i),\bar{c},
\label{eq:cond_dropout}
\end{equation}
where $\bar{c}$ is a fallback (zero depth, identity pose embedding, null/mean RGB). We set $\omega_i\in[0,1]$ as
\begin{equation}
\omega_i = v_i \cdot \exp!\left(-\frac{R_i}{\tau_R+\epsilon}\right),
\label{eq:omega_dropout}
\end{equation}
with normalized reprojection error $R_i$ and visibility ratio $v_i$.

\begin{figure*}[t]
    \centering
    \includegraphics[width=\linewidth]{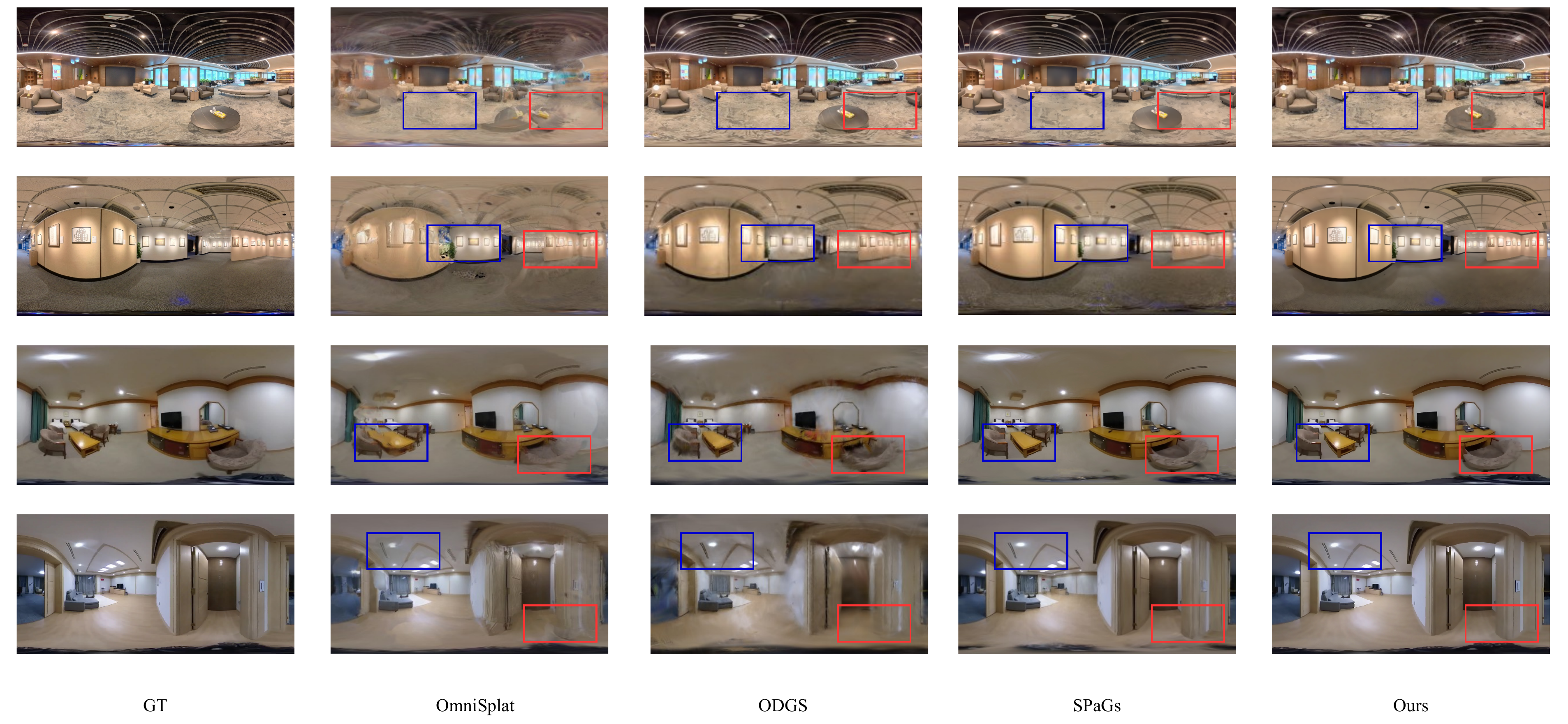}
    \caption{
    Qualitative comparison on Omniscenes and 360roam.
    From left to right: GT, OmniSplat, ODGS, SPaGS and \textbf{Ours}
    }
    \label{fig:qualitative_comparison}
\end{figure*}

\begin{table*}[t]
\centering
\caption[Visible-view quantitative comparison]{Quantitative comparison on visible views. Best results are highlighted in bold.}
\label{tab:visible_comparison}

\setlength{\tabcolsep}{6pt}
\renewcommand{\arraystretch}{1.2}

\begin{tabular}{l c c c c c}
\toprule
Dataset & MVSplat & OmniSplat & ODGS & SPaGs & Ours \\
\midrule
OmniScene 
& 19.89 / 0.662 / 0.261
& 23.78 / 0.784 / 0.285
& 24.41 / 0.810 / 0.217
& 25.67 / 0.890 / 0.318
& \textbf{30.86} / \textbf{0.893} / \textbf{0.148} \\

360Roam 
& 18.43 / 0.683 / 0.477
& 20.17 / 0.726 / 0.412
& 19.43 / 0.749 / 0.351
& 22.56 / 0.749 / 0.315
& \textbf{25.27} / \textbf{0.813} / \textbf{0.213} \\

\bottomrule
\end{tabular}
\label{tab:visible_comparison}
\end{table*}

\begin{table*}[t]
\caption{Ablation study on visible views.
Best results are highlighted in bold.}
\label{tab:ablation_all_metrics}
\centering
\small
\setlength{\tabcolsep}{5pt}
\renewcommand{\arraystretch}{1.15}

\begin{tabular}{l|ccccc}
\toprule
\multicolumn{1}{l|}{Scene} & $L_{c}+L_{d}$ & $L_{c}+L_{d}+L_{g}$ & $L_{c}+L_{d}+L_{a}$ & $L_{c}+L_{d}+L_{g}+L_{a}$ & w/o Diffusion \\
\midrule
room2  & 35.54 / 0.949 / 0.066 & 35.81 / 0.951 / 0.064 & 35.98 / 0.952 / 0.062 & \textbf{36.43 / 0.955 / 0.062} & 35.52 / 0.951 / 0.068 \\
room1  & 35.03 / 0.945 / 0.070 & 35.17 / 0.946 / 0.069 & 35.39 / 0.948 / 0.067 & \textbf{35.70 / 0.950 / 0.064} & 34.78 / 0.946 / 0.070 \\
office0& 33.40 / 0.934 / 0.083 & 33.63 / 0.935 / 0.081 & 33.82 / 0.937 / 0.079 & \textbf{34.42 / 0.941 / 0.077} & 33.58 / 0.937 / 0.084 \\
office2& 28.37 / 0.899 / 0.123 & 28.43 / 0.899 / 0.123 & 28.86 / 0.902 / 0.119 & \textbf{28.91 / 0.901 / 0.110} & 27.95 / 0.898 / 0.124 \\
office3& 32.27 / 0.926 / 0.092 & 32.46 / 0.927 / 0.090 & 32.86 / 0.930 / 0.087 & \textbf{33.07 / 0.931 / 0.088} & 32.25 / 0.927 / 0.098 \\
office4& 30.05 / 0.910 / 0.110 & 30.68 / 0.915 / 0.105 & 30.61 / 0.914 / 0.105 & \textbf{31.04 / 0.917 / 0.108} & 30.19 / 0.913 / 0.108 \\
\bottomrule
\end{tabular}

\end{table*}

\subsection{Depth-Constrained Gaussian Optimization}

Our objective targets the geometric instability of sparse panoramic reconstruction.
Given captured views, a feed-forward predictor provides initial pose and depth priors, which often exhibit heteroscedastic noise under large viewpoint gaps.
We therefore optimize 3D Gaussians with (i) a photometric data term on captured measurements,
(ii) a depth anchoring term,
(iii) a multi-view geometric consistency constraint, and
(iv) a geometry-aware anti-floater regularizer.
Importantly, synthesized targets are not treated as hard measurements; they contribute as reliability-weighted soft photometric and depth priors.

\paragraph{Reliability Weight.}
For each target view $t$, we compute a scalar reliability weight $\omega_t\in[0,1]$ from geometric consistency:
\begin{equation}
\omega_t
=
\exp\!\left(
-\alpha \, \bar{R}_t
-\beta \, \bar{\Delta}_t
\right),
\end{equation}
where $\bar{R}_t$ is the mean reprojection residual aggregated over reference views, and
$\bar{\Delta}_t$ measures the depth disagreement among warped reference depths.
Larger reprojection/depth inconsistency yields smaller $\omega_t$.
Parameters $\alpha,\beta>0$ control sensitivity.

\paragraph{Photometric Term ($\mathcal{L}_c$).}
We fit rendered radiance to captured views, and impose a reliability-weighted consistency prior on synthesized targets.
Let $\mathcal{V}_{\text{cap}}$ and $\mathcal{V}_{\text{syn}}$ denote captured and synthesized view sets, respectively:
\begin{equation}
\mathcal{L}_c
=
\sum_{t\in\mathcal{V}_{\text{cap}}}\ell_{\text{pho}}(\hat{I}_t,I_t)
\;+\;
\lambda_{\text{syn}}
\sum_{t\in\mathcal{V}_{\text{syn}}}\omega_t\,
\ell_{\text{pho}}(\hat{I}_t,I_t),
\end{equation}
where $\lambda_{\text{syn}}$ scales the overall strength of the synthesized-view prior and
$\omega_t$ suppresses unreliable targets.

\paragraph{Depth Term ($\mathcal{L}_d$).}
We anchor rendered depth on captured views, and add a reliability-weighted depth prior on synthesized targets:
\begin{equation}
\mathcal{L}_d
=
\sum_{t\in\mathcal{V}_{\text{cap}}}\|\hat{D}_t-D_t\|_1
\;+\;
\lambda_{\text{d,syn}}
\sum_{t\in\mathcal{V}_{\text{syn}}}\omega_t\,
\|\hat{D}_t-\tilde{D}_t\|_1,
\end{equation}
where $D_t$ denotes the predictor depth on captured view $t$ and
$\tilde{D}_t$ is a proxy target depth obtained by warping/fusing reference-view depth signals into the target view (thus independent of synthesized RGB).
The synthesized-view term acts as a soft prior rather than a measurement.

\paragraph{Multi-view Geometric Consistency ($\mathcal{L}_g$).}
Beyond per-view depth anchoring, we enforce cross-view coherence via reprojection.
For each target view $t$, valid pixels from source views $\mathcal{S}(t)$ are back-projected to $\mathbb{R}^3$,
transformed into the target frame, and reprojected onto the target image plane.
A z-buffer retains only proximal surfaces, yielding an aggregated reprojected depth map $D^{\mathrm{proj}}_t$.
To suppress outliers from occlusions or pose noise, we measure the forward--backward reprojection residual of each 3D point $\mathbf{p}_r$:
\[
\phi(\mathbf{p}_r) = \bigl\| \mathbf{p}_r - \mathcal{T}_{t\rightarrow r}(\mathcal{T}_{r\rightarrow t}(\mathbf{p}_r)) \bigr\|_2 ,
\]
and define a confidence weight
\[
w(\mathbf{p}_r)=
\begin{cases}
\exp(-\phi(\mathbf{p}_r)), & \phi(\mathbf{p}_r)<\tau,\\
0, & \text{otherwise}.
\end{cases}
\]
We penalize discrepancies between rendered and reprojected depths using a Huber loss $\rho(\cdot)$ with parameter $\delta$:
\begin{equation}
\mathcal{L}_g
=
\sum_{t} \frac{1}{Z_t}
\sum_{\mathbf{p}_r \in \mathcal{W}_t}
w(\mathbf{p}_r)\,
\rho\!\left(
\hat{D}_t(\pi_t(\mathbf{p}_r)) -
D^{\mathrm{proj}}_t(\pi_t(\mathbf{p}_r))
\right),
\end{equation}
where $\mathcal{W}_t$ is the set of verified reprojected points and $Z_t$ is a normalization constant.

\paragraph{Geometry-Aware Anti-Floater Regularization ($\mathcal{L}_a$).}
Sparse-view optimization can produce high-opacity Gaussians at implausible depths.
We penalize them only where reliable geometry contradicts the current rendering.
Define
\begin{equation}
m_{\mathrm{susp}}^{(t)}(\mathbf{u})
=\mathbb{I}[D_t(\mathbf{u})>0]\,
\mathbb{I}[\Delta D_t(\mathbf{u})>\tau_d]\,
\mathbb{I}[\hat{\alpha}_t(\mathbf{u})>a_0],
\end{equation}
\noindent where $\Delta D_t(\mathbf{u})\triangleq|\hat{D}_t(\mathbf{u})-D_t(\mathbf{u})|$.
Then
\begin{equation}
\mathcal{L}_a
=
\sum_t\,
\mathbb{E}_{\mathbf{u}\sim\Omega_t}\!\left[
m_{\mathrm{susp}}^{(t)}(\mathbf{u})
\left(\lambda_{\alpha}\,\hat{\alpha}_t(\mathbf{u})+\lambda_{D}\,\Delta D_t(\mathbf{u})\right)
\right],
\end{equation}
which suppresses opacity and depth deviations only in contradicted regions, without eroding valid surfaces.

\paragraph{Overall Objective.}
The total loss $\mathcal{L}$ is a weighted combination of the individual terms:
\begin{equation}
\mathcal{L} = \lambda_{c} L_c + \lambda_{d} L_d + \lambda_{g} L_g + \lambda_{a} L_a,
\end{equation}
where all hyper-parameters are held constant across different scene benchmarks to demonstrate the generalizability of the framework.
\begin{figure*}[t]
    \centering
    \includegraphics[width=\linewidth]{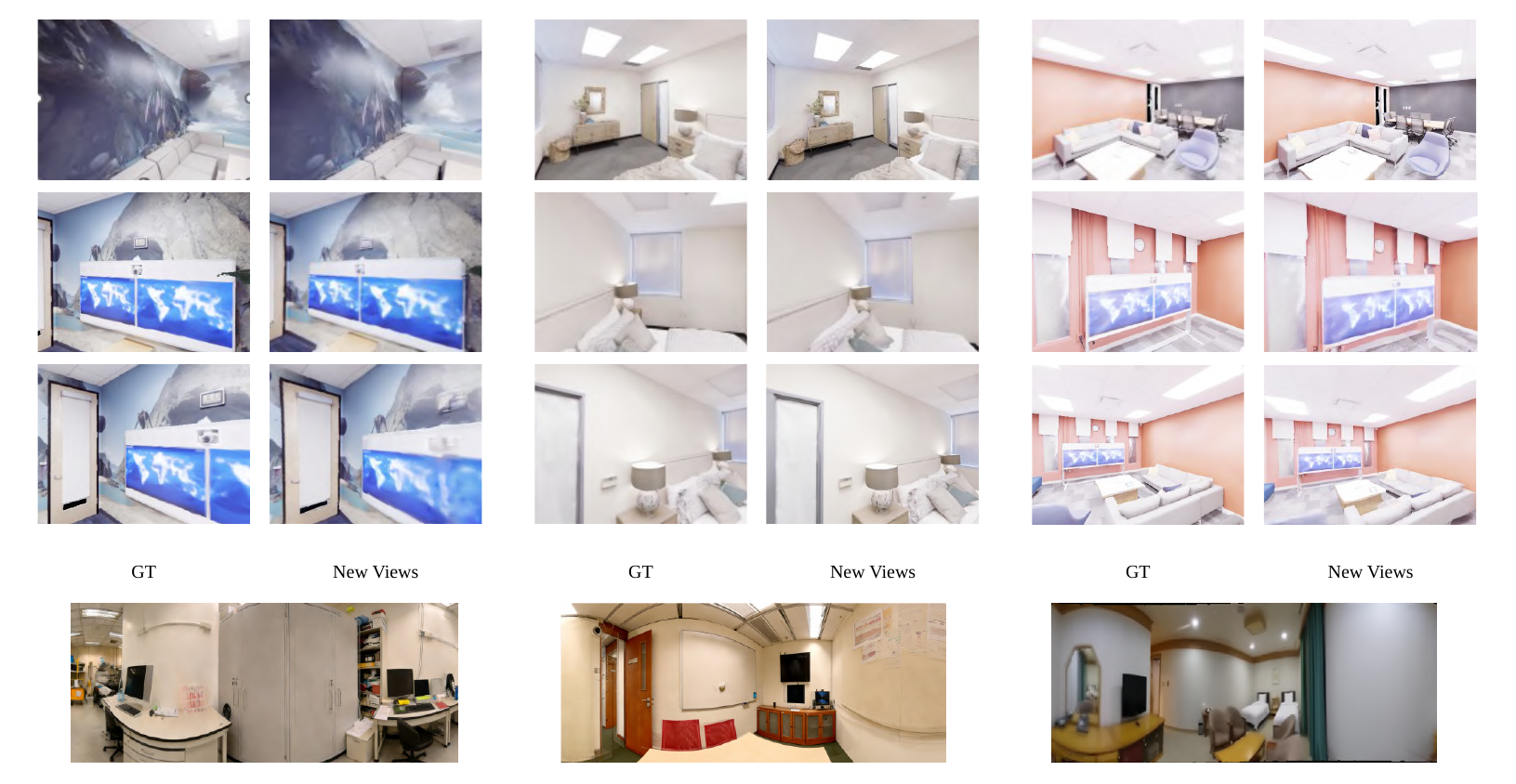}
    \caption{\textbf{We sample several local pinhole views from the panoramic view set and compare the diffusion-synthesized views with GT renderings at the same camera poses. In addition, we show synthesized panoramas on real-world scenes for qualitative evaluation of panoramic consistency and visual realism.}
}
    \label{fig:nvs}
\end{figure*}

\section{Experiments}

\subsection{Evaluation Protocol}
We evaluate PanoImager on \textbf{reconstruction fidelity} and \textbf{geometric consistency} under viewpoint extrapolation. And we report visible-view image metrics and, for non-visible regions, additional geometry-oriented proxies (depth reprojection error and free-space alignment) motivated by downstream mapping use cases.

\paragraph{Baselines and Metrics.} 
Experiments use \textbf{Replica}~\cite{straub2019replica} for geometric ground truth, and \textbf{OmniScenes}~\cite{wei2025omni} and \textbf{360Roam}~\cite{huang2022360roam} for real-world sparse-view tests. We compare against iterative and feed-forward baselines using PSNR, SSIM, and LPIPS to assess structural reliability.

\paragraph{Computational Efficiency.}
Unless otherwise stated, all experiments run on a single NVIDIA RTX A6000 (48GB). View completion takes $\sim$3 minutes per scene (resolution $W{\times}H$, $N_v{=}18$, $N{=}16$) to generate pose-aligned appearance priors, while 3DGS fusion is performed as offline or asynchronous background optimization for $K$ iterations. Due to the current compute/memory footprint, PanoImager is intended for map refinement or on-demand fallback rather than latency-critical onboard perception.

\subsection{Visible View Evaluation}
We qualitatively evaluate PanoImager against recent dense reconstruction methods, including OmniSplat, ODGS, and SPaGS. For a fair comparison, all methods are evaluated under the same sparse panoramic input budget (identical $N_p$ and the same selected frames), with matched rendering resolution and identical 3DGS optimization schedule (iterations and learning rates) where applicable. As shown in Fig.~\ref{fig:qualitative_comparison}, although these baselines perform well with dense inputs, they suffer from severe blur and depth misalignment in the sparse panoramic regime. In contrast, PanoImager maintains superior geometric fidelity and cross-view consistency.

Quantitative results in Tab.~\ref{tab:visible_comparison} further confirm our robustness to input sparsity. PanoImager consistently outperforms all baselines on OmniScenes and 360Roam. The performance gap is most pronounced on slender structures and complex textures, where limited overlap causes prior methods to produce view-dependent inconsistencies. By leveraging generative priors as an appearance regularizer under a fixed geometric scaffold, our approach better preserves fine-grained details from minimal inputs.
\subsection{Non-Visible View Evaluation}

We evaluate beyond-coverage synthesis on Replica with a low-overlap stress test. Since Replica is spatially compact, we focus on \emph{non-visible} targets rather than long-horizon extrapolation. Specifically, we select target views whose warp coverage from any input view is below $35\%$ of target pixels, ensuring minimal direct evidence from the observations.
For each scene, we use $N_p{=}4$ sparse panoramas and evaluate targets with near-zero overlap to the inputs, making the task extrapolative rather than interpolative. We render all methods with ground-truth poses to \emph{fix camera geometry} and enable a controlled comparison of completion quality.

As shown in Tab.~\ref{tab:diffusion_nonvisible},PanoImager achieves stable PSNR/SSIM/LPIPS across low-overlap non-visible targets, indicating that the proposed geometry-conditioned completion can provide useful appearance support even when direct observations are limited. Beyond appearance metrics, we additionally report geometry-oriented proxies in these low-overlap targets. The multi-view depth reprojection error $E_{\text{reproj}}$ is 0.32 m, and the Free-space IoU reaches 0.76 under ground-truth poses, suggesting improved geometric consistency in beyond-coverage completion.
\begin{table}[h]
\centering
\small
\setlength{\tabcolsep}{6pt}
\renewcommand{\arraystretch}{1.15}

\caption{Non-visible results on the Replica dataset.}
\label{tab:diffusion_nonvisible}

\begin{tabular}{lccc}
\hline
Scene   & PSNR   & SSIM   & LPIPS  \\
\hline
office0 & 17.13  & 0.538  & 0.349  \\
office2 & 17.78  & 0.578  & 0.321  \\
office3 & 18.61  & 0.655  & 0.279  \\
office4 & 17.83  & 0.568  & 0.329  \\
room0   & 18.77  & 0.678  & 0.249  \\
room1   & 21.63  & 0.758  & 0.199  \\
room2   & 19.85  & 0.738  & 0.225  \\
\hline
\end{tabular}

\end{table}
\subsection{Ablation Study and Performance Analysis}

\paragraph{Ablation on Visible Views.}
To verify the contribution of each objective term, we conduct an ablation study on Replica under a \emph{near-collinear straight-line} trajectory. In panoramic sensing, such motion induces weak parallax and ill-conditioned multi-view constraints, serving as a rigorous stress test for reconstruction robustness. 
We compare our full objective against several variants (Tab.~\ref{tab:ablation_all_metrics}), including a \emph{w/o Diffusion} baseline that optimizes 3DGS using only captured views. This isolates the effect of our generative densification, which populates the observation space to increase effective viewpoint overlap.

Our analysis reveals that: (i) $L_g$ (geometric consistency) alone may degrade performance in degenerate regimes due to the propagation of structured reprojection errors; (ii) $L_a$ (anti-floater regularization) improves robustness by pruning depth-inconsistent Gaussians but cannot rectify global misalignment; and (iii) removing diffusion consistently leads to inferior results, confirming that synthesized views are essential for stabilizing multi-view constraints. When combined with diffusion-based densification, $L_g$ and $L_a$ become highly complementary: $L_g$ enforces structural alignment across the expanded observation set, while $L_a$ filters spurious primitives, yielding the best overall fidelity.

\paragraph{Pose Estimation Stability.}
In the near-collinear translation setting, traditional SLAM/SfM methods are prone to failure due to ill-conditioned triangulation under weak parallax, leading to brittle initialization and tracking. By contrast, our feed-forward initialization remains stable, achieving an average \textbf{ATE} of \textbf{0.362\,m} with \textbf{RPE$_t$} \textbf{0.138\,m} and \textbf{RPE$_r$} \textbf{0.781$^\circ$/m}, providing a reliable geometric scaffold for downstream Gaussian optimization.

\paragraph{Discussion on Generative-Geometric Coupling.}
Diffusion-only synthesis can produce visually plausible views, but it often lacks the cross-view consistency required for 3D reconstruction, leading to structural drift during extrapolation. By coupling generative completion with depth-guided Gaussian reconstruction, PanoImager anchors the generative prior to a coherent 3D representation, suppressing floating artifacts and improving robustness in sparse panoramic mapping.

\subsection{Results and Analysis}
The results show that diffusion-only synthesis can produce visually plausible images, but often lacks cross-view geometric consistency, especially under large viewpoint extrapolation. Such errors typically appear as structural misalignment and view-dependent artifacts.
By coupling diffusion-based view generation with depth-guided Gaussian reconstruction, our method improves both geometric stability and appearance consistency in sparse panoramic settings. The reconstructed Gaussian representation anchors the generative prior to a more coherent 3D structure, providing a more reliable basis for novel-view synthesis.
The visible-view ablations further indicate that diffusion guidance, depth-constrained optimization, and anti-floater regularization play complementary roles. Together, they improve reconstruction quality, suppress floating artifacts, and enhance robustness under sparse panoramic inputs.

\section{Limitations and Future Work}
PanoImager prioritizes structural robustness over strict metric precision in extremely sparse regimes where purely geometric pipelines often fail. This choice comes with trade-offs: (i) Uncertainty: under extreme extrapolation, elevated aleatoric uncertainty may yield visually plausible completions that do not guarantee metric accuracy; (ii) Seams: discrete tangent-space charting can introduce mild artifacts in low-overlap regions; and (iii) Efficiency: the current computational cost limits use to asynchronous background mapping or on-demand fallback refinement rather than real-time onboard perception. Future work will incorporate explicit uncertainty quantification for safety-critical navigation and model distillation to narrow the deployment gap.

\section*{Acknowledgement}
This work was funded by JSPS KAKENHI, grant numbers JP24K21173 and JP24H00351. This work was also supported by the UTokyo-SMBC Forest GX project.

\bibliographystyle{IEEEtran}
\bibliography{reference}

\end{document}